\pdfoutput=1
\documentclass{article}

\usepackage{PRIMEarxiv}

\usepackage[utf8]{inputenc}
\usepackage[T1]{fontenc}
\usepackage{microtype}
\usepackage[full]{textcomp}
\usepackage{hyperref}

\usepackage{booktabs,tabularx,array}

\usepackage{amsmath, amsthm}
\usepackage{mathtools}
\usepackage{nicefrac}
\usepackage{thmtools}
\usepackage{cleveref}

\usepackage{stix2}
\usepackage{inconsolata}

\usepackage{graphicx}
\usepackage{adjustbox}
\usepackage{xcolor}
\usepackage{xspace}
\usepackage{caption}
\usepackage{natbib}
\usepackage[section]{placeins}

\title{Iris Content Guard: Conversation Review at Scale}
\author{\textbf{Anaelia Ovalle} \quad \textbf{Drew Proud} \quad \textbf{Mart Bakler} \quad \textbf{Ben Batten} \\ \textbf{Christopher Grimm} \quad \textbf{Shomik Jain} \quad \textbf{Aymen Kallala} \quad \textbf{Nick Fram} \\[0.5em] Reflection}
\begin{document}
\maketitle
\begin{abstract}
Teams that review logged conversations between users and AI assistants need a classifier that catches policy violations without flooding reviewers with safe content, at a cost low enough to run over large volumes of logs. We introduce Iris Content Guard (ICG), an open-weight safety classifier that scores user requests and assistant responses across 14 harm categories in a single forward pass, reading no policy prompt and generating no text. We built ICG by pruning an in-house mixture-of-experts language model to 18 layers (28.7B total parameters, 3.2B active per token) and fine-tuning it on labeled conversations and soft targets distilled from a larger teacher model. Because ICG outputs continuous scores rather than written verdicts, teams can set a threshold to match their false-positive budget. On our Iris Safety Evaluations (ISE) benchmark, ICG achieves higher F1 than all seven open safeguard and general-purpose models we tested, reaching 88.8\% on requests and 89.0\% on responses while using about 2 to 14 times less compute per item. When each model may flag at most 5\% of safe items, ICG catches 78.6\% of harmful requests and 93.2\% of harmful responses, ahead of 66.1\% and 65.7\% for the next best models; prompted reasoning models drop sharply under this limit because their verdict probabilities cluster near 0 or 1. Across ten public safety benchmarks filtered for agreement with our policy, ICG leads all specialized classifiers and matches much larger reasoning models at a fraction of their compute. We release ICG's weights under the Apache 2.0 license.
\end{abstract}

\FloatBarrier
\section{Introduction}
\label{sec:1}

Reviewing logged conversations between users and AI assistants requires identifying policy violations without sending safe content for unnecessary review. Missed violations leave harmful content undetected, while false positives consume reviewer time. Because conversation logs are large, the classifier also has to be cheap enough to score every request and response.

Teams can choose between two kinds of open models for this job. Dedicated safeguard models classify content for safety: earlier models such as Llama Guard \citep{inan2023llamaguard}, ShieldGemma \citep{zeng2024shieldgemma}, WildGuard \citep{han2024wildguard}, and Qwen3Guard \citep{zhao2025qwen3guard} ship with their own safety taxonomies, while newer models such as GPT-OSS-Safeguard \citep{openai2025safeguard} and Shieldstral \citep{mistral2026shieldstral} follow a policy written in the prompt. Teams can also prompt general-purpose language models with a policy and ask for a verdict \citep{zheng2023judging}. Both prompted approaches process a long policy prompt and generate output tokens, with recent general-purpose and safeguard reasoning models generating a chain of thought before giving a verdict. That extra processing raises inference cost even when teams cache and reuse the policy prompt across items. A written verdict also gives a team no direct way to trade false positives against detection. Even when teams derive scores from a reasoning model's verdict probabilities, those scores tend to cluster near 0 or 1, so raising the threshold to meet a low false-positive target discards many violations.

We introduce Iris Content Guard (ICG), an open-weight safety classifier for conversations between users and AI assistants. ICG takes a user request, optionally with the corresponding assistant response, and returns 14 scores between 0 and 1 in a single forward pass, one for each harm category in our policy. Because ICG learns the policy during training, it reads no policy prompt and writes no text. We built ICG by pruning layers and experts from an in-house mixture-of-experts language model so only a small fraction of parameters run per token, then fine-tuning it on jury-labeled data and soft targets from a larger teacher model. Returning continuous scores rather than a yes-or-no verdict lets teams set a threshold that matches their review capacity.

On our Iris Safety Evaluations (ISE) benchmark, ICG achieves higher F1 than all seven open safeguard and general-purpose models we tested, on both requests (88.8\%) and responses (89.0\%), while the baselines use about 2 to 14 times ICG's compute per item. When each model may flag at most 5\% of safe items, ICG catches 78.6\% of harmful requests and 93.2\% of harmful responses; the next best models catch 66.1\% (Qwen3.8-27B) and 65.7\% (Shieldstral-1.0-3B). The models closest to ICG in F1, DeepSeek-V4.1-Flash on requests and GLM-5.3-Flash on responses, use about 5 to 10 times ICG's compute per item and catch only 46.6\% of harmful requests and 48.3\% of harmful responses, respectively, at the same limit. On public safety benchmarks, ICG is the strongest specialized classifier after we filter each benchmark for agreement with our policy, and the strongest open model on class-balanced versions of those subsets. Many models exceed 95\% on several of these benchmarks, suggesting they are approaching saturation.

Our contributions are:
\begin{itemize}
    \item \textbf{A sparse single-pass classifier.} A layer- and expert-pruned mixture-of-experts model with dedicated request and response heads that return 14 independent category scores in a single forward pass (\hyperref[sec:2]{Section 2} and \hyperref[sec:A]{Appendix A}).
    \item \textbf{A training recipe for overlapping categories.} Selective ``far-zero'' negative assignment for harm categories that often co-occur, combined with soft targets distilled from a larger teacher model (\hyperref[sec:2]{Section 2} and \hyperref[sec:A]{Appendix A}).
    \item \textbf{An evaluation of accuracy and cost.} A comparison against seven open safeguard and general-purpose models on ISE and ten public benchmark sets, covering F1, detection at a 5\% false-positive limit, detection by harm category, and inference compute (\hyperref[sec:3]{Section 3} and \hyperref[sec:B]{Appendix B}).
\end{itemize}
We release ICG's weights under the Apache 2.0 license, with links in \hyperref[sec:4]{Section 4}.

\FloatBarrier
\section{How Iris Content Guard works}
\label{sec:2}

We built Iris Content Guard from a language model we developed entirely in-house, from pretraining through post-training: a 32-layer transformer with 3 dense layers followed by 29 mixture-of-experts layers \citep{shazeer2017moe,fedus2021switch}, each with 256 experts and 8 active per token. To choose how many layers to keep, we fit linear probes \citep{alain2016probes} across the frozen parent model to predict safety violations at each depth. Probe performance plateaus around layer 16 and peaks before the final layers, with layer 18 outperforming the full 32-layer model on both requests and responses \citep[cf.][]{gromov2024deeper}. We therefore kept the first 18 layers (3 dense and 15 mixture-of-experts layers) and pruned each mixture-of-experts layer to the 128 experts routed most frequently on a calibration sample \citep[cf.][]{lu2024experts}. The resulting classifier has 28.7 billion parameters, activates 3.2 billion per token, and runs on a single 80 GB H100 GPU.

In place of a language-modeling head, we added two linear classification heads: one for user requests and one for assistant responses. Each head reads the final hidden state at the last token of its target message, with prior conversation turns kept as context, and outputs 14 independent scores between 0 and 1, one per harm category. Categories are scored independently, so one item can be flagged in several at once.

We fine-tuned the backbone and both heads jointly on 60,361 records labeled by a jury of LLM judges \citep{verga2024juries} and 106,978 unlabeled records supervised with soft probabilities from an ensemble of 52-layer teacher models \citep{hinton2015distilling}. On jury-labeled records, treating every unlabeled category on a harmful item as negative would penalize categories that often co-occur (such as fraud and illegal activities) and overwhelm rare categories with negatives. Instead, we marked safe records negative in all 14 categories, assigned each harmful record at most one explicit negative (a ``far zero'') on an unrelated category, and masked remaining unlabeled categories from the loss, capping each category's far zeros at its positive count. Full architecture, data, pruning, and optimization details appear in \hyperref[sec:A]{Appendix A}.

\FloatBarrier
\section{How ICG compares with open models}
\label{sec:3}

\FloatBarrier
\subsection{Classification performance and cost}
\label{sec:3.1}

We compare ICG with open models that review conversations for harmful content: dedicated safeguard models (GPT-OSS-Safeguard-20B and GPT-OSS-Safeguard-120B \citep{openai2025safeguard}, Shieldstral-1.0-3B \citep{mistral2026shieldstral}, and Qwen3Guard-Gen-8B \citep{zhao2025qwen3guard}) and general-purpose models (DeepSeek-V4.1-Flash \citep{deepseek2026v41flash}, GLM-5.3-Flash \citep{zai2026glm53flash}, and Qwen3.8-27B \citep{qwen2026qwen38}). Models that accept policy instructions receive ICG's safety policy. We report F1 separately for user requests and assistant responses, using each model's generated verdict or default score threshold. ICG flags an item when any of its 14 category scores is at least 0.5.

\begin{figure}[htbp]
\centering
\includegraphics[width=\linewidth,height=0.4\textheight,keepaspectratio]{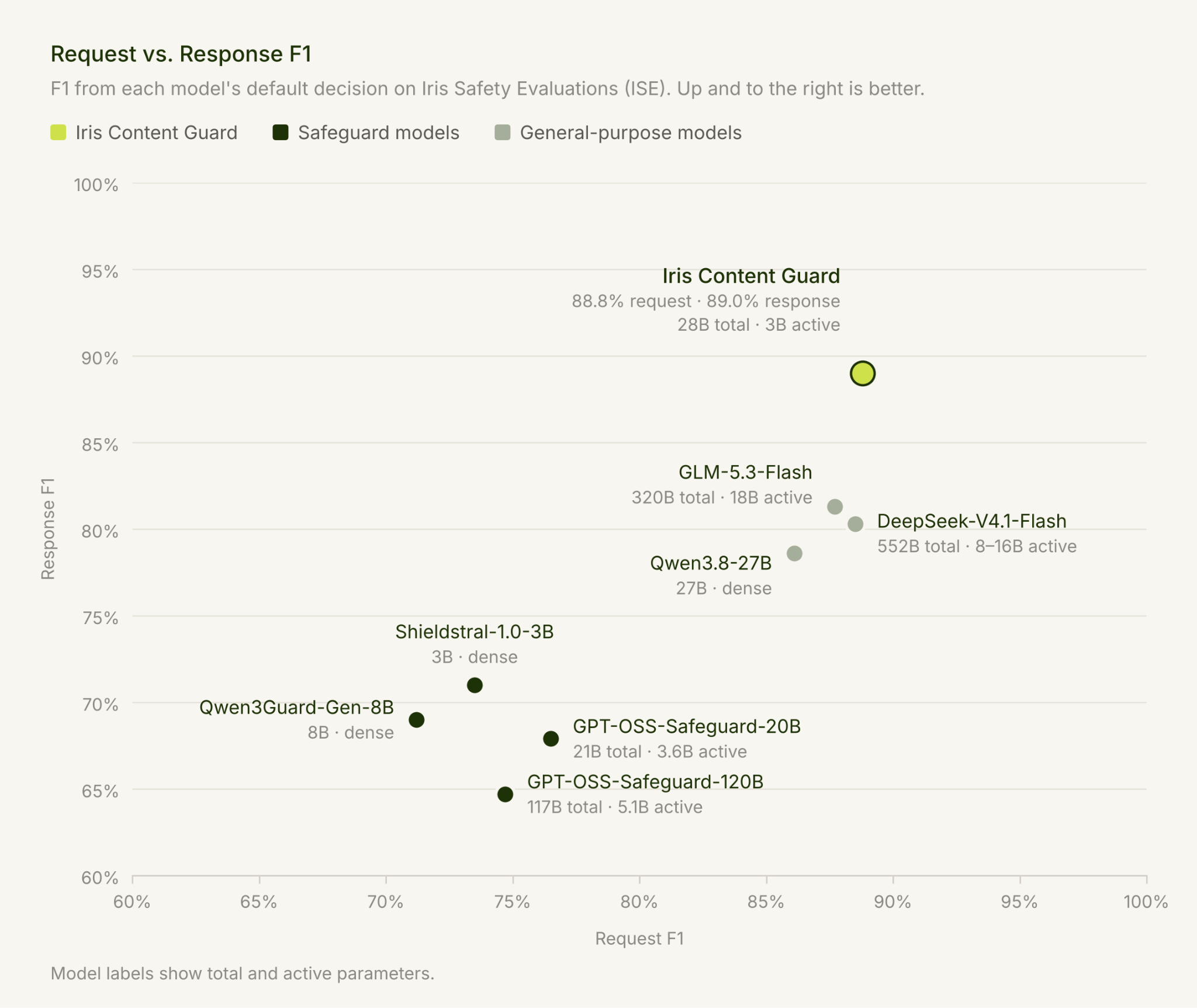}
\phantomsection\label{fig:1}
\caption*{Figure 1. Request F1 (x-axis) and response F1 (y-axis) on ISE, using each model's generated verdict or default score threshold. Models toward the upper right are more accurate on both. Reasoning models run at their highest reasoning-effort setting. Qwen3Guard-Gen-8B is plotted as the average of its strict and loose modes.}
\end{figure}

ICG achieves the highest F1 of the models we evaluated, with 88.8\% on requests and 89.0\% on responses (\hyperref[fig:1]{Figure 1}). The dedicated safeguard baselines use about 2 to 3.2 times ICG's compute per item yet trail ICG by more than 10 F1 points on both requests and responses, while the general-purpose models come closer in F1 at about 5 to 14 times ICG's compute per item. DeepSeek-V4.1-Flash reaches 88.5\% on requests with roughly 19 times ICG's total parameters, while GLM-5.3-Flash reaches 81.3\% on responses with roughly 11 times as many.

The models achieve these F1 scores at different false-positive rates. Therefore, teams with limited reviewer capacity may adjust score thresholds to reduce unnecessary review. The next section compares how many violations each model catches when it may flag no more than 5\% of safe items. \hyperref[sec:B]{Appendix B} gives full results, model settings, and compute calculations.

\FloatBarrier
\subsection{Detection under a false-positive limit}
\label{sec:3.2}

To compare models under a fixed review budget, we select for each model the lowest score threshold that flags no more than 5\% of safe items on ISE, separately for requests and responses. ICG uses the maximum of its 14 category scores, while for the prompted models we derive a score from the probabilities assigned to their verdict labels. These scores let teams choose a cutoff based on their tolerance for false positives, rather than relying only on the generated verdict.

\begin{figure}[htbp]
\centering
\includegraphics[width=\linewidth,height=0.4\textheight,keepaspectratio]{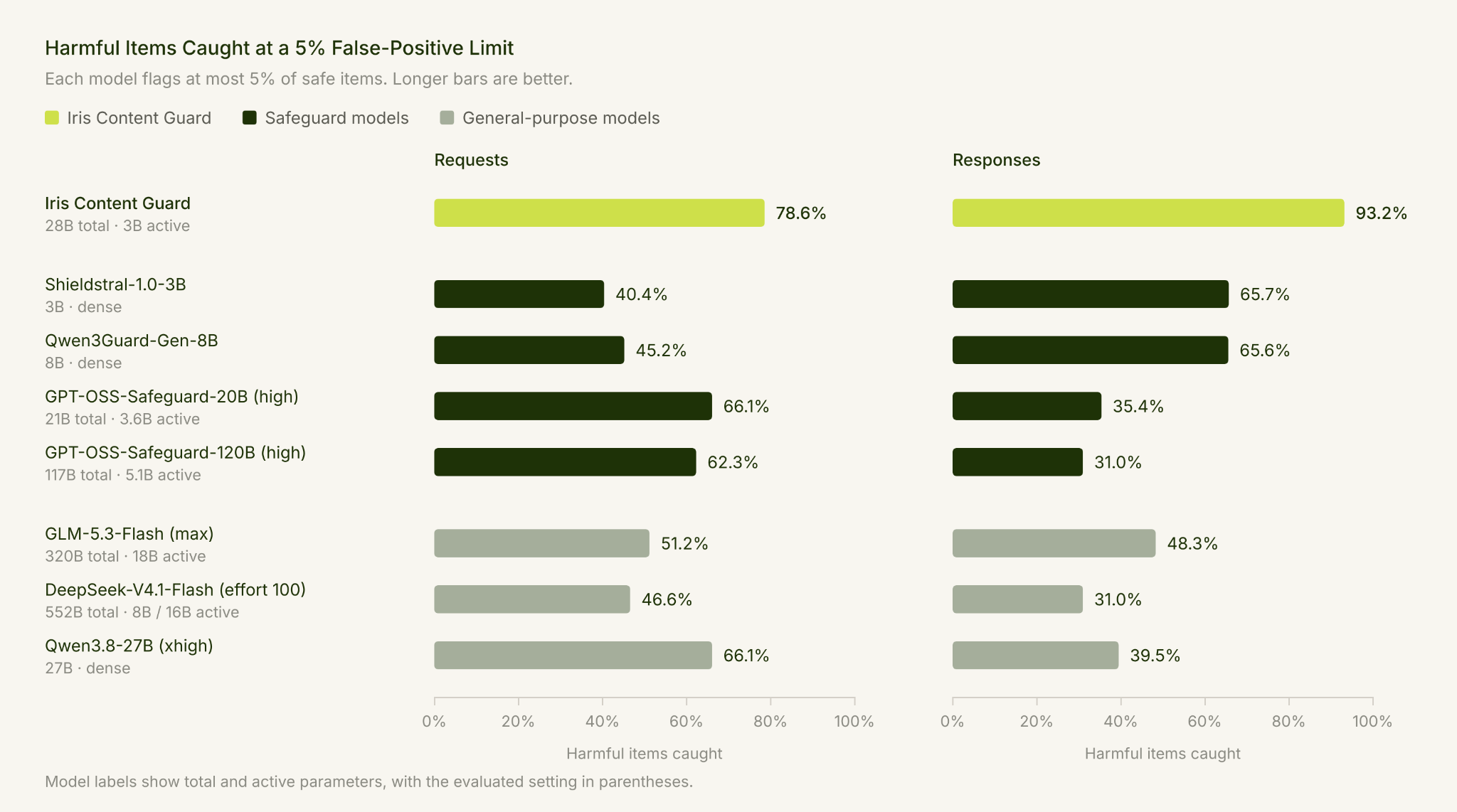}
\phantomsection\label{fig:2}
\caption*{Figure 2. Harmful requests and responses caught when each model's threshold is set to flag at most 5\% of safe items, separately for requests and responses. For Qwen3Guard, we average results from the strict and loose classification modes.}
\end{figure}

At this 5\% limit, ICG catches 78.6\% of harmful requests and 93.2\% of harmful responses (\hyperref[fig:2]{Figure 2}). The next best models catch 66.1\% of harmful requests (Qwen3.8-27B) and 65.7\% of harmful responses (Shieldstral-1.0-3B).

DeepSeek and GLM came closest to ICG in the F1 comparison, but neither is the strongest competitor at the 5\% limit. DeepSeek nearly matches ICG's request F1, at 88.5\% versus 88.8\%, but catches 46.6\% of harmful requests at this limit, compared with ICG's 78.6\%. GLM has the second-highest response F1, at 81.3\%, yet catches 48.3\% of harmful responses at the same limit, compared with ICG's 93.2\%. Meeting a low false-positive target requires scores that effectively separate violations from safe items the model might otherwise flag. Reasoning model scores tend to cluster near 0 or 1, failing to separate these cases, so raising their threshold to hit the 5\% limit discards many violations. Because ICG is explicitly trained to output scores, raising its threshold drops far fewer violations.

\begin{figure}[htbp]
\centering
\includegraphics[width=\linewidth,height=0.4\textheight,keepaspectratio]{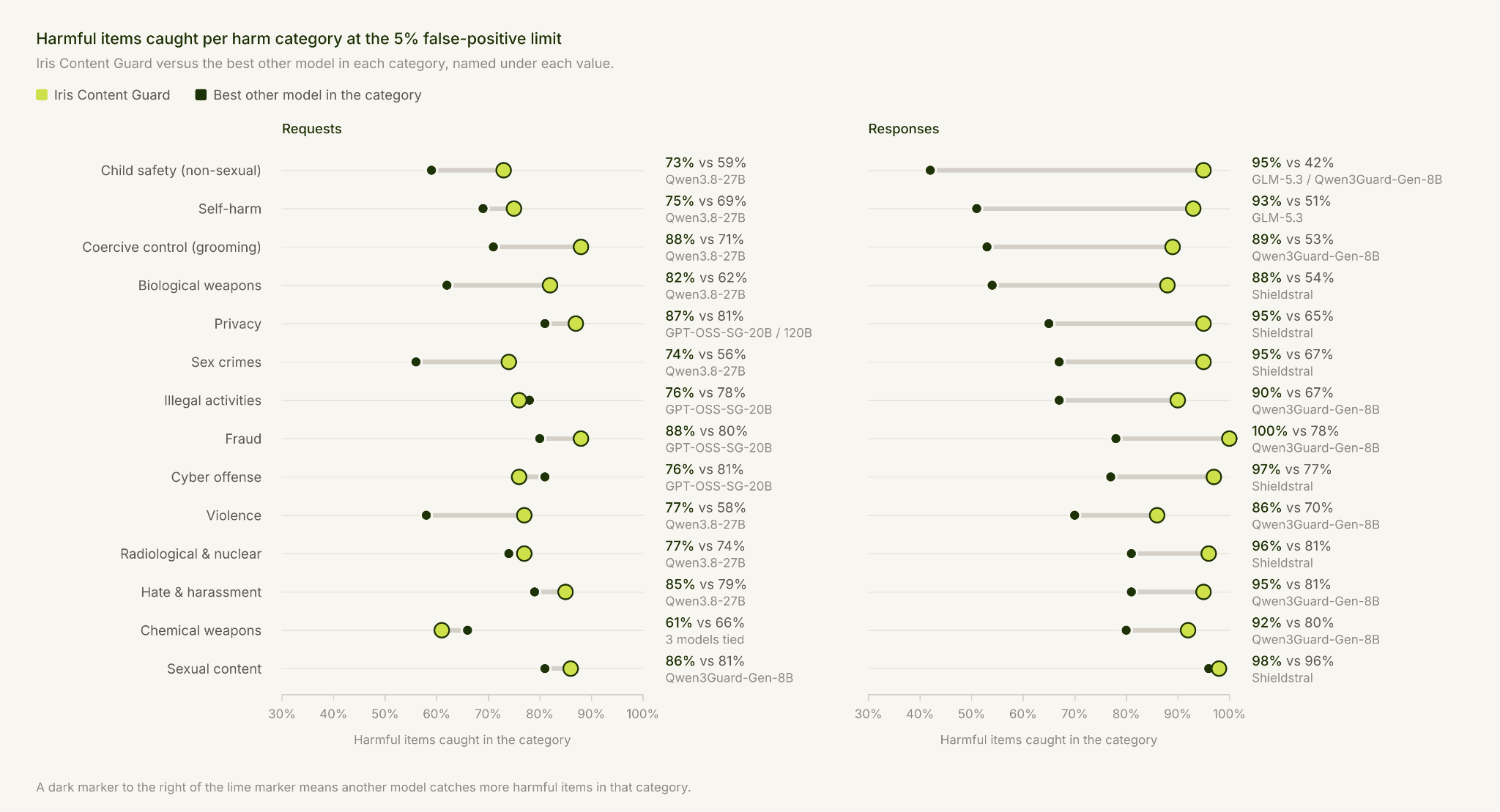}
\phantomsection\label{fig:3}
\caption*{Figure 3. Harmful items caught by category, using the same overall request and response thresholds as \hyperref[fig:2]{Figure 2}. Light green is ICG, dark green is the best other model in each category. Qwen3Guard results are averaged over its strict and loose modes in both figures. Detailed category results are in \hyperref[tab:B.3.2]{Table B.3.2}.}
\end{figure}

\hyperref[fig:3]{Figure 3} breaks down detection by harm category, comparing ICG with the best other model in each category. ICG catches more harmful responses in all 14 categories. The largest gap is in non-sexual child safety, where ICG catches 94.9\% of harmful responses, compared with 42\% for the next best models, GLM-5.3-Flash and Qwen3Guard-Gen-8B (averaging Qwen3Guard's strict and loose modes). On requests, ICG leads in 11 of 14 categories but trails the best other model on chemical weapons (61.1\% vs. 65.6\%), cyber offense (76.3\% vs. 80.6\%), and illegal activities (76.4\% vs. 78.2\%).

\FloatBarrier
\subsection{Results on Public Benchmarks}
\label{sec:3.3}

To measure how Iris Content Guard performs beyond our internal datasets, we also evaluated it on ten public safety benchmark sets, covering four request sets (Aegis v2 \citep{ghosh2025aegis2}, HarmBench \citep{mazeika2024harmbench}, OpenAI Moderation \citep{markov2022holistic}, and WildGuardTest \citep{han2024wildguard}) and six response sets (Aegis v2, BeaverTails \citep{ji2023beavertails}, HarmBench, Qwen3GuardTest \citep{zhao2025qwen3guard}, WildGuardTest, and XSTestHarm \citep{rottger2023xstest}). Because each benchmark was built around its own safety policy, we report three views of each benchmark: the full set with its original labels, a policy-filtered subset consistent with our policy, and a class-balanced version of that filtered subset created by upsampling the minority label with replacement. To build the filtered subsets, DeepSeek-V4.1-Flash and GLM-5.3-Flash classified each item against our safety policy, and we removed the 16.0\% of requests and 16.2\% of responses where both judges agreed that the original label conflicted with our policy. Further details on the filtering process appear in \hyperref[sec:B.4]{Appendix B.4}.

\begin{figure}[htbp]
\centering
\includegraphics[width=\linewidth,height=0.4\textheight,keepaspectratio]{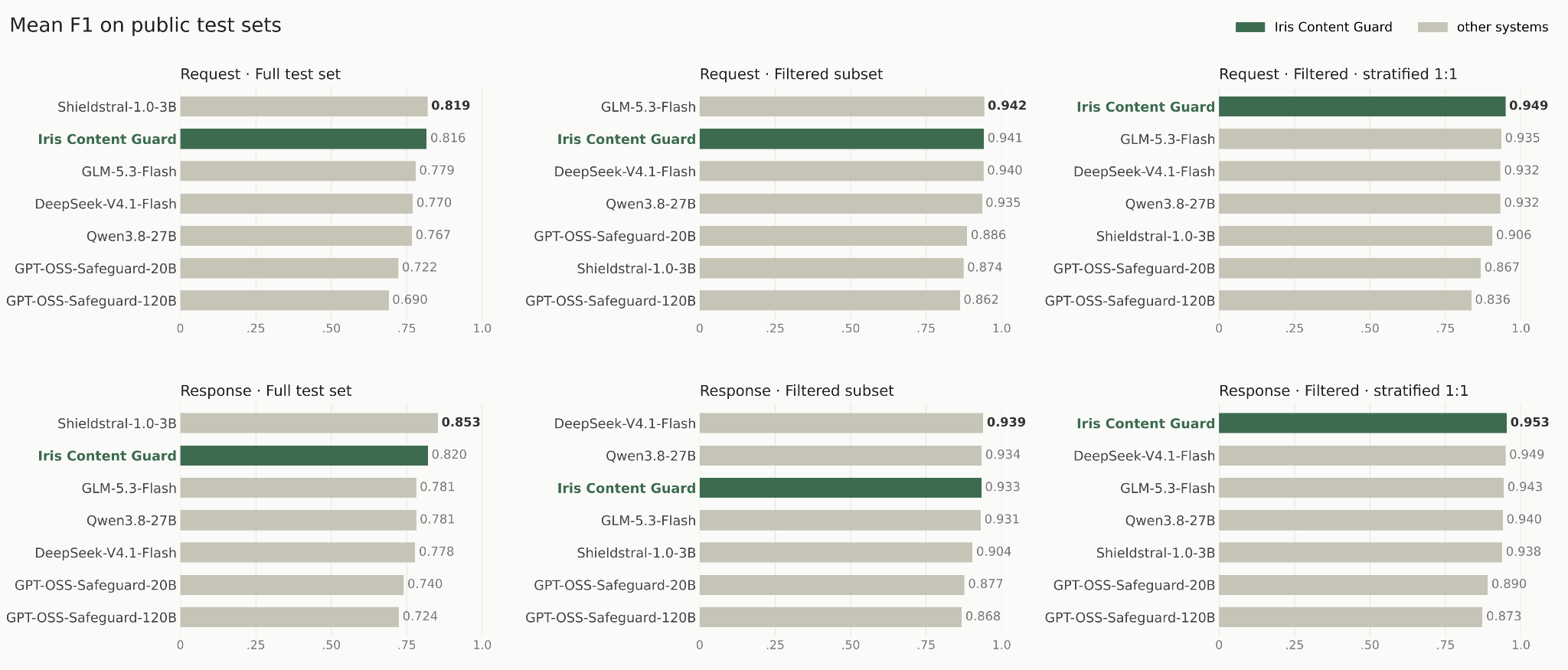}
\phantomsection\label{fig:4}
\caption*{Figure 4. Mean F1 on public benchmarks for requests (top) and responses (bottom), using each model's default decisions. Columns show the original test sets, policy-filtered subsets, and class-balanced versions of those subsets. Each benchmark receives equal weight.}
\end{figure}

\begin{figure}[htbp]
\centering
\includegraphics[width=\linewidth,height=0.4\textheight,keepaspectratio]{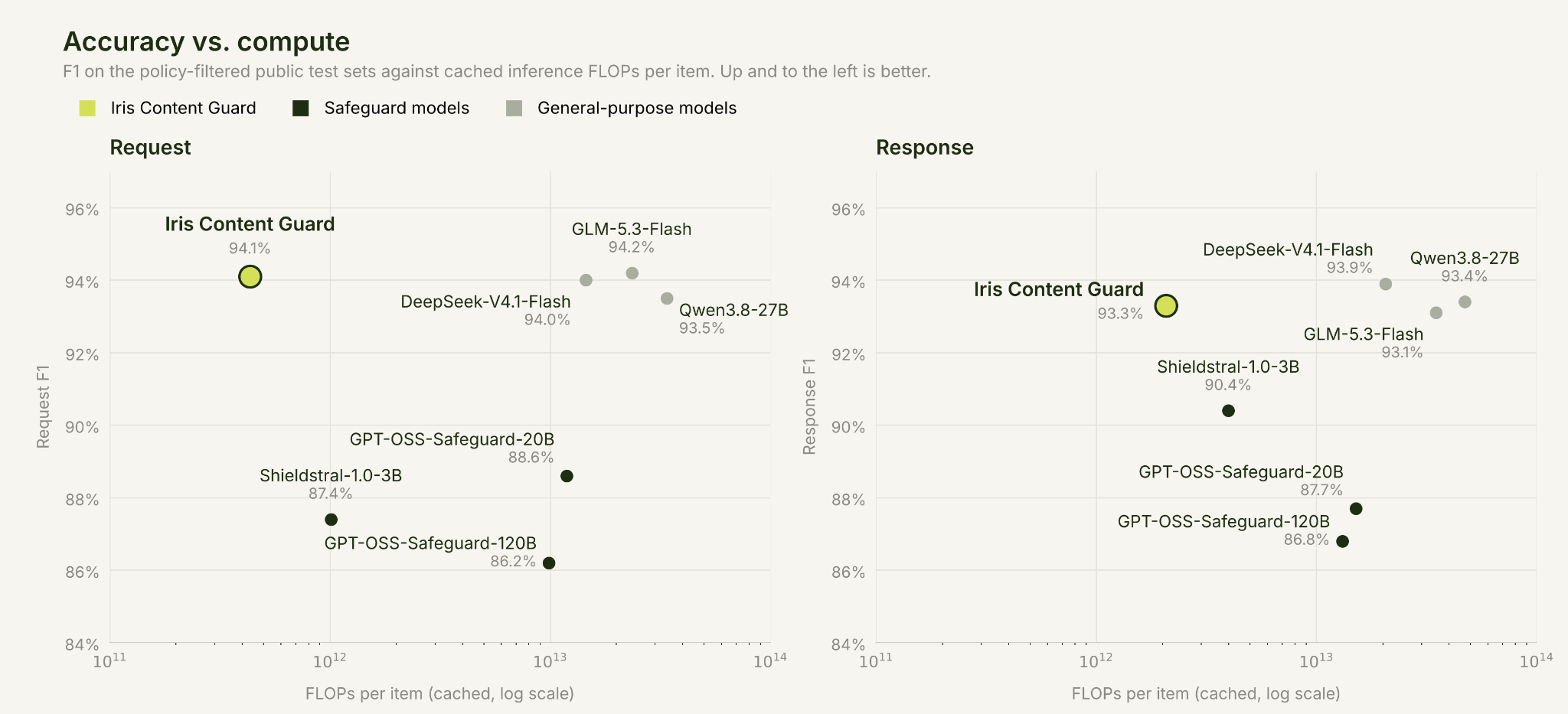}
\phantomsection\label{fig:5}
\caption*{Figure 5. Mean F1 on the policy-filtered public benchmarks versus inference compute per item, shown separately for requests (left) and responses (right). Points toward the upper left combine higher F1 with lower compute.}
\end{figure}

We evaluated ICG alongside the six baselines that accept a custom policy in the prompt; models with a fixed built-in policy, such as Qwen3Guard-Gen-8B, cannot take our policy for a like-for-like comparison. We report mean F1 at each model's default decision boundary (\hyperref[fig:4]{Figure 4}). On the policy-filtered subsets, Iris Content Guard is the strongest specialized classifier on both requests and responses. When we include the general-purpose reasoning models, ICG is slightly lower on mean F1 than the top reasoning model, but offers a clear Pareto improvement against inference compute per item (\hyperref[fig:5]{Figure 5}). On the class-balanced subsets, ICG achieves the highest mean F1 among all evaluated open models on both requests and responses, reflecting balanced classification across safe and harmful items.

Compared with ISE, both specialized and general-purpose models reach substantially higher scores on these public benchmarks, with many models exceeding 95\% F1 on several sets. This pattern suggests that these public benchmarks are approaching saturation and no longer provide enough signal to distinguish strong safety classifiers.

\FloatBarrier
\section{Using Iris Content Guard}
\label{sec:4}

ICG helps teams prioritize logged conversations for review against its 14-category safety policy. Teams can use the highest category score to prioritize conversations for general review, or use individual category scores to set different thresholds and review actions for different types of harm.

Teams should treat the default threshold of 0.5 as a starting point and choose thresholds on a labeled sample of their own conversations, since false-positive rates on production data may differ from those measured on ISE, and verify that the policy categories in \hyperref[sec:C]{Appendix C} cover their needs. ICG is intended for English conversational text between users and AI assistants.

Future work includes extending ICG beyond English to multilingual conversations and agentic transcripts with tool calls, as well as building harder gray-area evaluations as current public benchmarks saturate. Iris Content Guard's weights are available under the Apache 2.0 license on Hugging Face at \url{https://huggingface.co/reflection/Iris-Content-Guard}.

\FloatBarrier
\bibliographystyle{plainnat}
\bibliography{refs}

\FloatBarrier
\appendix

\FloatBarrier
\section{Training Details}
\label{sec:A}

\textbf{Model.} As summarized in \hyperref[sec:2]{Section 2}, the parent model has 3 dense transformer layers followed by 29 mixture-of-experts layers \citep{shazeer2017moe,fedus2021switch}, with 256 experts in each mixture-of-experts layer. ICG retains the first 18 layers (3 dense and 15 mixture-of-experts layers) and 128 experts per mixture-of-experts layer, with 8 experts active per token (28.7 billion total parameters, 3.2 billion active per token). The request and response linear classification heads each read the final hidden state at the last token of their target message, with the preceding conversation as context, and output 14 category scores.

\textbf{Training data.} The 60,361 binary-labeled records drew from public safety training sets, synthetically generated examples, and human-written data. A jury of LLM judges \citep{verga2024juries} classified requests and responses separately as safe or harmful under our safety policy and identified which categories harmful items violated. To assess label quality, we compared the jury labels to human annotators over 2,574 requests and 4,274 responses; the jury agreed with the majority label from 3 annotators on 88\% of requests and 83\% of responses.

For the jury-labeled records, each label becomes a per-category target: 1 for a violation and 0 for a non-violation, with unlabeled categories excluded from the loss. A safe record is negative in every category. Each harmful example received at most one explicit negative, a ``far zero'', on a category that is neither labeled nor related to a labeled one, where two categories are related if records often violate both. Across the dataset, no category received more far zeros than it has positives, so rare categories are not biased toward low scores.

\textbf{Teacher and distillation.} To supervise the additional 106,978 unlabeled records, we used a larger in-house teacher model \citep{hinton2015distilling}. This pool drew from WildGuardMix \citep{han2024wildguard}, WildChat \citep{zhao2024wildchat}, Aegis v2 \citep{ghosh2025aegis2}, WildJailbreak \citep{jiang2024wildteaming}, and OR-Bench \citep{cui2024orbench}, together with generated responses and internal prompts and responses. We deduplicated this pool against existing labeled data and public test sets.

The teacher is a 52-layer transformer with 3 dense layers followed by 49 mixture-of-experts layers, each containing 256 experts with 8 active per token. We fine-tuned it end to end on an earlier snapshot of 46,839 labeled records. For each category, we averaged its predicted probabilities across three training runs with different random seeds to obtain the teacher target.

We jointly trained ICG on the existing binary labels and soft teacher probabilities using binary cross-entropy, masking categories without a training target.

\begin{figure}[htbp]
\centering
\includegraphics[width=\linewidth,height=0.4\textheight,keepaspectratio]{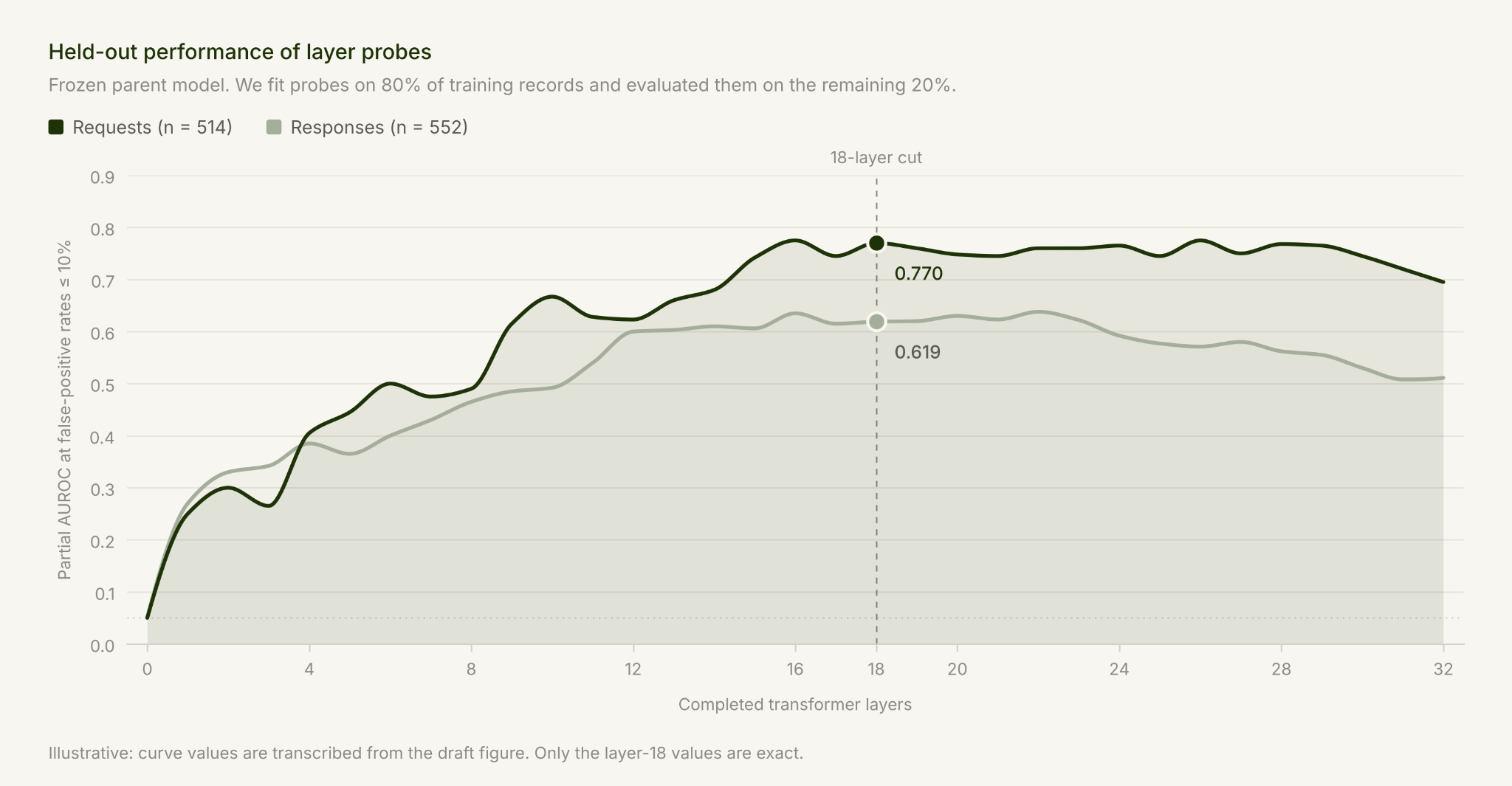}
\phantomsection\label{fig:A1}
\caption*{Figure A1. Linear-probe performance on held-out training records: 514 requests and 552 responses. The vertical line marks the retained depth of 18 layers. These results measure the frozen parent's representations before classifier fine-tuning.}
\end{figure}

\textbf{Layer selection.} When choosing which layers to prune, we fit separate logistic regression probes \citep{alain2016probes} for requests and responses at each frozen layer of the parent model, predicting whether an item violates any of the 14 safety categories. Using a sample of 3,894 records, split 80\% for training and 20\% for evaluation by record (2,570 records contain a request and 2,767 contain a response), we evaluated performance using partial ROC AUC (0--10\% FPR). Performance plateaus around layer 16, with response probes peaking between layers 16 and 23 (\hyperref[fig:A1]{Figure A1}). At layer 18, probes score 0.770 on requests and 0.619 on responses, compared to 0.696 and 0.512 after all 32 layers. Consequently, we retain the first 18 layers and remove the remaining 14 \citep[cf.][]{gromov2024deeper}.

\textbf{Expert pruning.} We select experts by how frequently the parent model routes tokens to them \citep[cf.][]{lu2024experts}, using a calibration sample of 34,100 examples (23.4M tokens) drawn from the training data and balanced across data sources. The parent divides each layer's 256 experts into eight groups of 32. We retain the 16 most frequently used experts in each group, leaving 128 experts per layer while preserving the routing structure. The retained experts receive 78\% of routed tokens on average across the 15 expert layers of the 18-layer model (66\% at minimum), compared with 50\% if routing were uniform. We apply expert pruning to the full 32-layer parent, then load its first 18 layers for fine-tuning.

\textbf{Optimization.} We fully fine-tune the backbone and both heads with Adam \citep{kingma2015adam}, using $\beta_1 = 0.9$, $\beta_2 = 0.999$, $\epsilon = 10^{-8}$, a batch size of 256, and no weight decay. The learning rate warms up for 25 steps to $3 \times 10^{-5}$, then follows cosine decay to $3 \times 10^{-6}$. We train for two epochs over the prepared training collection, corresponding to 2,614 optimizer steps. Training uses a context window of 32k tokens, although the base model supports up to 128k. To check that ICG can be used beyond its training length, we test it on conversations of up to 128k tokens and find no degradation in its scores.

\FloatBarrier
\section{Evaluation Details and Additional Results}
\label{sec:B}

\FloatBarrier
\subsection{Evaluation Setup}
\label{sec:B.1}

We evaluate on two sources: Iris Safety Evaluations (ISE), which is our main evaluation, and a set of public safety benchmarks.

\FloatBarrier
\subsubsection{Iris Safety Evaluations (ISE) v1}
\label{sec:B.1.1}

Iris Safety Evaluations (ISE) v1 covers 15 harm categories, 14 of which are included in our evaluation. ISE prompts are used directly as requests and are categorized into four control layers: C1 (clearly harmless), C2 (gray area but harmless), C3 (gray area but harmful), and C4 (clearly harmful). We group C1--C2 as non-violating and C3--C4 as violating.

Responses are generated for both violating and non-violating requests using a combination of abliterated \citep{arditi2024refusal} and non-abliterated models. Each response is evaluated and assigned a label based on a majority vote from a panel of three LLM judges, which are distinct from the models used to label the ICG training data (\hyperref[sec:A]{Appendix A}). To ensure diversity, the labeled responses are filtered for quality and coherence, retaining both complete and partial answers. This process yields a final dataset comprising both violating and non-violating responses.

For multi-turn conversations, each turn is evaluated as an individual example retaining prior turns as context; only the final request or response is scored, with responses always paired alongside their corresponding request.

In total, the evaluation dataset consists of 7,533 scored items with 3,195 user requests (1,479 harmful and 1,716 safe) and 4,338 assistant responses (1,040 harmful and 3,298 safe). Every item is assigned a single harm category used for the category-level breakdowns in \hyperref[fig:3]{Figure 3} and \hyperref[tab:B.3.2]{Table B.3.2}. All items are English conversations between a user and an AI assistant, without tool calls or multi-step agent actions. We do not evaluate ICG on other languages or on agentic transcripts.

\FloatBarrier
\subsubsection{Baselines}
\label{sec:B.1.2}

\textbf{Models.} We compare ICG against 7 open models. Three are general-purpose models that we prompt with our policy: DeepSeek-V4.1-Flash \citep{deepseek2026v41flash}, GLM-5.3-Flash \citep{zai2026glm53flash}, and Qwen3.8-27B \citep{qwen2026qwen38}. Three are safeguard models trained specifically to apply a policy supplied in the prompt, which also receive our policy: GPT-OSS-Safeguard-20B, GPT-OSS-Safeguard-120B \citep{openai2025safeguard}, and Shieldstral-1.0-3B \citep{mistral2026shieldstral}. The seventh model, Qwen3Guard-Gen-8B \citep{zhao2025qwen3guard}, is a safeguard model with its own built-in policy, which we include as a reference point for ISE evaluations. Qwen3Guard outputs Safe, Controversial, or Unsafe, so we evaluate it under both \textit{strict} (Controversial and Unsafe count as violating) and \textit{loose} (only Unsafe counts as violating) mappings.

\textbf{Prompting and configuration.} Every baseline except Qwen3Guard-Gen-8B receives our 14-category policy. For the safeguard models that accept a policy, we follow each model's native instructions for policy placement and prompt formatting. For general-purpose models, we place the policy in the system prompt and state the classification task in both the system and user messages in order to help the model distinguish the conversation being evaluated from instructions to the model. ICG receives no policy at inference time. In all main figures, DeepSeek-V4.1-Flash, GLM-5.3-Flash, Qwen3.8-27B, and GPT-OSS-Safeguard run at their highest reasoning effort; Shieldstral-1.0-3B and Qwen3Guard-Gen-8B have no effort setting. Results at lower-effort settings appear in \hyperref[tab:B.2.2]{Table B.2.2}.

\FloatBarrier
\subsection{Classification Performance and Compute}
\label{sec:B.2}

We compare models in two ways: at each model's default decision and at a score threshold chosen to meet a false-positive limit. In both, an item on which a model returns no usable answer counts as wrong: a missed harmful item, or a false flag on a safe one.

\FloatBarrier
\subsubsection{Performance at Default Decisions}
\label{sec:B.2.1}

In \hyperref[fig:1]{Figure 1} and \hyperref[tab:B.2.1]{Table B.2.1}, each model uses its default decision. ICG flags an item when its highest category score is at least 0.5, and Shieldstral-1.0-3B uses the same 0.5 threshold. The reasoning models are scored on their generated verdict, and Qwen3Guard-Gen-8B on its generated label under each mapping. We compute F1 separately for requests and responses, with harmful as the positive class. Because these default decisions flag different shares of safe items, \hyperref[tab:B.2.1]{Table B.2.1} reports both F1 and the share of safe items each model flags.

\begin{table}[htbp]
\centering
\small
\phantomsection\label{tab:B.2.1}
\caption*{Table B.2.1. F1 and safe items flagged at each model's default decision on ISE, with model size and mean compute per item relative to ICG (\hyperref[tab:B.2.2]{Table B.2.2}).}
\begin{adjustbox}{max width=\linewidth}
\begin{tabular}{lcccccc}
\toprule
\textbf{Model} & \textbf{Parameters (total / active per token)} & \textbf{Compute per item ($\times$ ICG)} & \textbf{Request F1} & \textbf{Response F1} & \textbf{Safe requests flagged} & \textbf{Safe responses flagged} \\
\midrule
Iris Content Guard & 28.7B / 3.2B & 1.0$\times$ & 88.8\% & 89.0\% & 11.6\% (199) & 5.8\% (191) \\
\midrule
\multicolumn{7}{l}{\textit{Safeguard models}} \\
Shieldstral-1.0-3B & 3B / all & 2.0$\times$ & 73.5\% & 71.0\% & 17.4\% (298) & 4.5\% (148) \\
Qwen3Guard-Gen-8B, strict & 8B / all & 2.8$\times$ & 78.4\% & 73.9\% & 29.3\% (503) & 6.4\% (212) \\
Qwen3Guard-Gen-8B, loose & 8B / all & 2.8$\times$ & 64.0\% & 64.1\% & 5.9\% (101) & 1.7\% (55) \\
GPT-OSS-Safeguard-20B (high) & 21B / 3.6B & 3.0$\times$ & 76.5\% & 67.9\% & 4.8\% (82) & 10.1\% (333) \\
GPT-OSS-Safeguard-120B (high) & 117B / 5.1B & 3.2$\times$ & 74.7\% & 64.7\% & 5.1\% (88) & 8.6\% (283) \\
\midrule
\multicolumn{7}{l}{\textit{General-purpose models}} \\
GLM-5.3-Flash (max) & 320B / 18B & 10.0$\times$ & 87.7\% & 81.3\% & 11.1\% (191) & 11.0\% (364) \\
DeepSeek-V4.1-Flash (effort 100) & 552B / 8B / 16B & 5.3$\times$ & 88.5\% & 80.3\% & 8.6\% (148) & 12.8\% (421) \\
Qwen3.8-27B (xhigh) & 27B / all & 14.2$\times$ & 86.1\% & 78.6\% & 7.7\% (132) & 12.2\% (402) \\
\bottomrule
\end{tabular}
\end{adjustbox}
\end{table}

\FloatBarrier
\subsubsection{Inference Compute}
\label{sec:B.2.2}

We calculate inference compute in TFLOP ($10^{12}$ floating-point operations). The main parameter-dependent term is approximately $2 \times (\text{active parameters per token}) \times (\text{tokens processed})$. We account separately for prompt processing, token generation, and output projection, and include attention operations. Active-parameter counts come from model cards where available, and from configurations or checkpoint tensor shapes otherwise. DeepSeek-V4.1-Flash's model card gives 8B active parameters for tokens it reads and 16B for tokens it writes, and we use both; for GLM-5.3-Flash we use the card's 18B. For DeepSeek-V4.1-Flash's total, we use the card's 552B backbone parameters; the checkpoint also holds a 196B conditional-memory module that is read by lookup rather than computed per token. Token counts come from the same evaluation runs used to measure classification performance.

Our main comparison assumes each prompted model's fixed policy prefix is reused across items. We calculate inference compute for each item and report the mean across requests and responses, giving each included item equal weight. Compute ratios are relative to ICG's combined mean. External compute figures are lower bounds because retry tokens were not recorded. Shieldstral generates a one-token verdict, so nearly all of its compute comes from processing the input.

The lower-effort runs use less compute, but none matches ICG's detection rate on either requests or responses at the 5\% false-positive limit (\hyperref[tab:B.2.2]{Table B.2.2}). GPT-OSS-Safeguard's low- and high-effort runs used different serving builds, so their comparison does not isolate the effect of reasoning effort.

\begin{table}[htbp]
\centering
\small
\phantomsection\label{tab:B.2.2}
\caption*{Table B.2.2. Mean inference compute per item and harmful items caught at the 5\% false-positive limit, for the main and lower-effort settings. Compute assumes policy-prefix reuse and is averaged across requests and responses, with ratios relative to ICG. Detection is reported separately for requests and responses. Qwen3Guard results average its strict and loose modes. FLOP calculations include output layers. ICG's 3.2B active parameters include its input embedding table.}
\begin{adjustbox}{max width=\linewidth}
\begin{tabular}{lcccccc}
\toprule
\textbf{Model (effort)} & \textbf{Active params / token} & \textbf{Tokens written (mean)} & \textbf{Compute (TFLOP)} & \textbf{$\times$ ICG} & \textbf{Requests caught @5\%} & \textbf{Responses caught @5\%} \\
\midrule
Iris Content Guard & 3.2B & 0 & 9.8 & 1.0$\times$ & 78.6\% & 93.2\% \\
Shieldstral-1.0-3B & 3.43B & 1 & 19.9 & 2.0$\times$ & 40.4\% & 65.7\% \\
Qwen3Guard-Gen-8B & 7.57B & 12 & 27.4 & 2.8$\times$ & 45.2\% & 65.6\% \\
GPT-OSS-Safeguard-20B (high) & 3.6B & 1,758 & 29.9 & 3.0$\times$ & 66.1\% & 35.4\% \\
GPT-OSS-Safeguard-20B (low) & 3.6B & 146 & 14.6 & 1.5$\times$ & 40.6\% & 26.1\% \\
GPT-OSS-Safeguard-120B (high) & 5.1B & 875 & 31.5 & 3.2$\times$ & 62.3\% & 31.0\% \\
GPT-OSS-Safeguard-120B (low) & 5.1B & 125 & 21.5 & 2.2$\times$ & 40.3\% & 19.8\% \\
DeepSeek-V4.1-Flash (effort 100) & 8B / 16B & 627 & 52.5 & 5.3$\times$ & 46.6\% & 31.0\% \\
DeepSeek-V4.1-Flash (effort 50) & 8B / 16B & 334 & 42.3 & 4.3$\times$ & 55.1\% & 38.6\% \\
GLM-5.3-Flash (max) & 18B & 930 & 98.4 & 10.0$\times$ & 51.2\% & 48.3\% \\
GLM-5.3-Flash (low) & 18B & 20 & 64.2 & 6.5$\times$ & 41.2\% & 39.6\% \\
Qwen3.8-27B (xhigh) & 25.6B & 766 & 140.3 & 14.2$\times$ & 66.1\% & 39.5\% \\
Qwen3.8-27B (low) & 25.6B & 554 & 127.5 & 13.0$\times$ & 70.6\% & 57.2\% \\
\bottomrule
\end{tabular}
\end{adjustbox}
\end{table}

\begin{figure}[htbp]
\centering
\includegraphics[width=\linewidth,height=0.4\textheight,keepaspectratio]{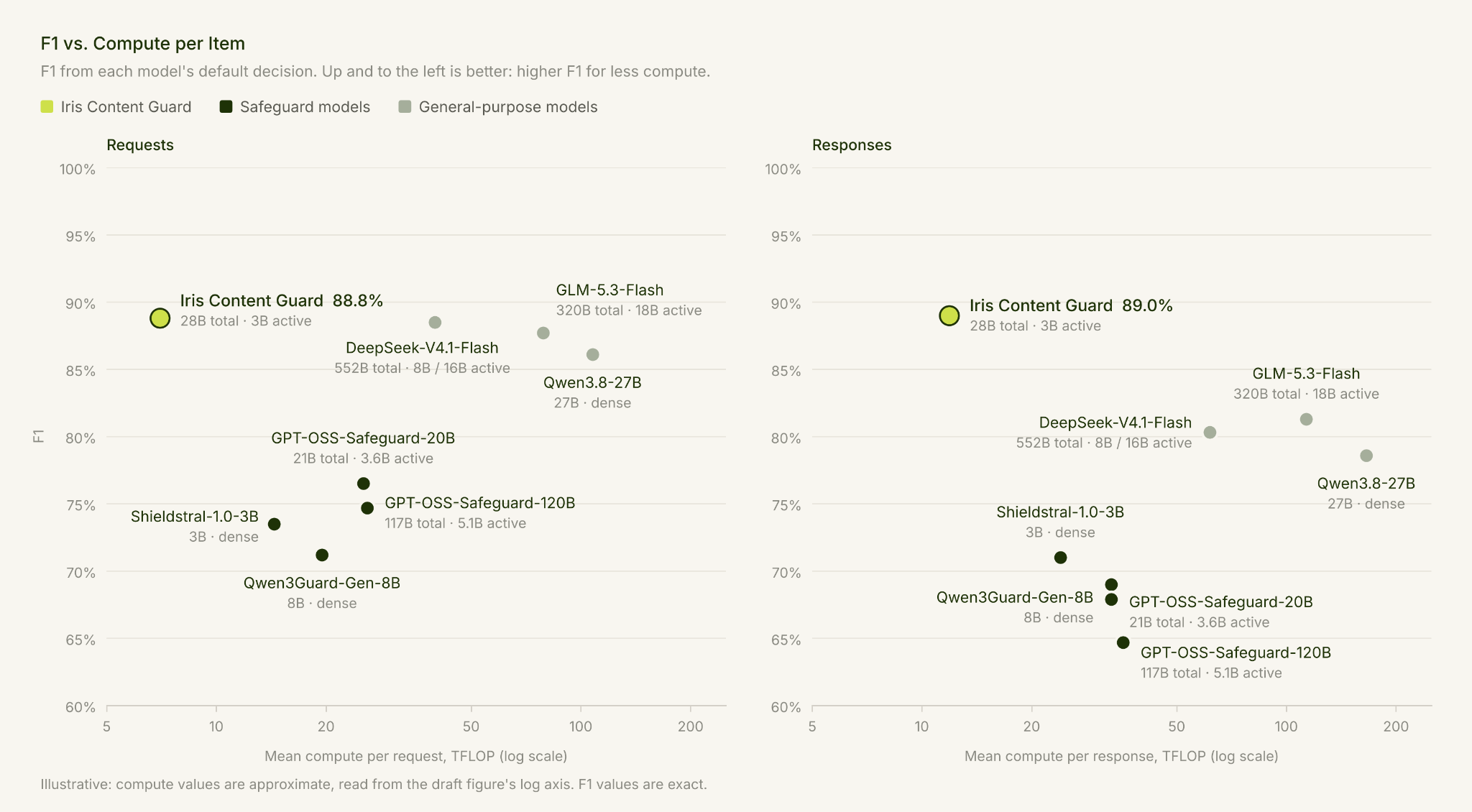}
\phantomsection\label{fig:B.2.2}
\caption*{Figure B.2.2. F1 at each model's default decision against mean compute per request (left) and per response (right), log scale. Compute assumes each prompted model's fixed policy prefix is reused across items; ICG has no policy prompt.}
\end{figure}

\FloatBarrier
\subsection{Detection Under a False-Positive Limit}
\label{sec:B.3}

\FloatBarrier
\subsubsection{Scores and Thresholds}
\label{sec:B.3.1}

This comparison needs a continuous score for each model. ICG's score is the maximum of its 14 category scores. Scores for the baselines are derived from log-probabilities at the first verdict or label token:

\begin{itemize}
  \item \textbf{DeepSeek, GLM, Qwen3.8, GPT-OSS-Safeguard, Shieldstral:} the probability assigned to tokens the parser accepts as a violating verdict, divided by the total probability assigned to accepted violating and safe verdict tokens.
  \item \textbf{Qwen3Guard:} probabilities are taken from the first tokens of its three labels. The strict score is the combined Controversial and Unsafe probability, normalized over all three labels; the loose score uses only the Unsafe probability in the numerator.
\end{itemize}

For each model, we use the lowest score threshold that flags no more than 5\% of safe items, chosen separately for requests and responses; items with tied scores are kept together. The 10\% comparison uses the same rule with a 10\% limit. Because each threshold is chosen on the same items it is scored on, we also repeated the comparison with 5-fold cross-validation grouped by conversation, over 20 random splits. No model's detection at the 5\% limit changed by more than 1.2 points, and ICG caught more harmful items than every other model in every split. The reasoning models' scores tend to sit close to 0 or 1. When such a model's default decision already flags more than the limit (\hyperref[tab:B.2.1]{Table B.2.1}), meeting the limit requires a much stricter threshold, and its detection at the limit can fall well below its detection at the default decision.

\begin{table}[htbp]
\centering
\small
\phantomsection\label{tab:B.3.1}
\caption*{Table B.3.1. Harmful items caught when each model flags no more than 5\% or 10\% of safe items, and partial ROC area over false-positive rates of 0--10\% (normalized to 0--1), on ISE. Each model's threshold is the lowest score that flags no more than the stated share of safe items, chosen separately for requests and responses. Figures 2 and 3 average Qwen3Guard's strict and loose modes; on its own, the strict mode catches 66.3\% of harmful responses at the 5\% limit.}
\begin{adjustbox}{max width=\linewidth}
\begin{tabular}{lcccccc}
\toprule
\textbf{Model / configuration} & \textbf{Caught @5\% requests} & \textbf{Caught @5\% responses} & \textbf{Caught @10\% requests} & \textbf{Caught @10\% responses} & \textbf{pAUROC requests} & \textbf{pAUROC responses} \\
\midrule
Iris Content Guard & 78.6\% (1,162) & 93.2\% (969) & 88.6\% (1,310) & 97.5\% (1,014) & 0.728 & 0.879 \\
\midrule
\multicolumn{7}{l}{\textit{Safeguard models}} \\
Shieldstral-1.0-3B & 40.4\% (598) & 65.7\% (683) & 55.3\% (818) & 76.3\% (794) & 0.378 & 0.591 \\
Qwen3Guard-Gen-8B, strict & 43.2\% (639) & 66.3\% (690) & 57.9\% (856) & 78.1\% (812) & 0.372 & 0.624 \\
Qwen3Guard-Gen-8B, loose & 47.1\% (697) & 64.9\% (675) & 61.9\% (916) & 77.0\% (801) & 0.435 & 0.609 \\
GPT-OSS-Safeguard-20B (high) & 66.1\% (977) & 35.4\% (368) & 74.9\% (1,108) & 67.4\% (701) & 0.549 & 0.348 \\
GPT-OSS-Safeguard-120B (high) & 62.3\% (922) & 31.0\% (322) & 70.4\% (1,041) & 63.7\% (663) & 0.517 & 0.339 \\
\midrule
\multicolumn{7}{l}{\textit{General-purpose models}} \\
GLM-5.3-Flash (max) & 51.2\% (757) & 48.3\% (502) & 80.5\% (1,191) & 87.0\% (905) & 0.476 & 0.474 \\
DeepSeek-V4.1-Flash (effort 100) & 46.6\% (689) & 31.0\% (322) & 87.3\% (1,291) & 78.6\% (817) & 0.479 & 0.342 \\
Qwen3.8-27B (xhigh) & 66.1\% (978) & 39.5\% (411) & 84.4\% (1,248) & 76.2\% (793) & 0.598 & 0.392 \\
\bottomrule
\end{tabular}
\end{adjustbox}
\end{table}

\FloatBarrier
\subsubsection{Harm Detection by Category}
\label{sec:B.3.2}

\hyperref[tab:B.3.2]{Table B.3.2} breaks detection down by harm category, with every model held to the same 5\% limit. ``ICG overall'' counts an item as caught when any of ICG's 14 category scores clears the cutoff; ``ICG own head'' counts it only when the score for the item's labeled category does, which shows whether ICG catches harm under the right category. The own head alone catches 75.3\% of harmful requests and 91.5\% of harmful responses, close to ICG's overall 78.6\% and 93.2\%. The largest gaps are in sexual-content and illegal-activities requests, which overlap with neighboring categories such as sex crimes and fraud. Even counting only its own head, ICG catches more than the best other model in 9 of 14 request categories and 13 of 14 response categories.

\begin{table}[htbp]
\centering
\small
\phantomsection\label{tab:B.3.2}
\caption*{Table B.3.2. Harmful items caught per harm category at the 5\% false-positive limit, using the same cutoffs as \hyperref[tab:B.3.1]{Table B.3.1}. ``ICG overall'' uses ICG's highest category score; ``ICG own head'' uses only the score for the item's labeled category, at the same cutoff. ``Best other model'' is the highest of the eight other configurations in each category. Ties are named. This table lists Qwen3Guard's strict and loose modes separately, while \hyperref[fig:3]{Figure 3} averages them, so a few best-other values differ from the figure (for example, 43.0\% here versus 42\% for non-sexual child safety responses).}
\begin{adjustbox}{max width=\linewidth}
\begin{tabular}{lccp{0.3\linewidth}}
\toprule
\textbf{Category (harmful items)} & \textbf{ICG overall} & \textbf{ICG own head} & \textbf{Best other model} \\
\midrule
\multicolumn{4}{l}{\textit{Requests}} \\
Biological weapons (130) & 81.5\% (106) & 81.5\% (106) & 62.3\% (81), Qwen3.8-27B \\
Chemical weapons (157) & 61.1\% (96) & 59.2\% (93) & 65.6\% (103), GLM-5.3-Flash / Qwen3.8-27B / GPT-OSS-Safeguard-120B \\
Child safety (non-sexual) (90) & 73.3\% (66) & 68.9\% (62) & 58.9\% (53), Qwen3.8-27B \\
Coercive control (grooming) (120) & 88.3\% (106) & 82.5\% (99) & 70.8\% (85), Qwen3.8-27B \\
Cyber offense (93) & 76.3\% (71) & 75.3\% (70) & 80.6\% (75), GPT-OSS-Safeguard-20B \\
Fraud (88) & 87.5\% (77) & 84.1\% (74) & 79.5\% (70), GPT-OSS-Safeguard-20B \\
Hate \& harassment (113) & 85.0\% (96) & 83.2\% (94) & 78.8\% (89), Qwen3.8-27B \\
Illegal activities (110) & 76.4\% (84) & 66.4\% (73) & 78.2\% (86), GPT-OSS-Safeguard-20B \\
Privacy (112) & 86.6\% (97) & 86.6\% (97) & 81.2\% (91), GPT-OSS-Safeguard-20B / GPT-OSS-Safeguard-120B \\
Radiological \& nuclear (94) & 76.6\% (72) & 74.5\% (70) & 74.5\% (70), Qwen3.8-27B \\
Self-harm (93) & 75.3\% (70) & 74.2\% (69) & 68.8\% (64), Qwen3.8-27B \\
Sex crimes (87) & 73.6\% (64) & 72.4\% (63) & 56.3\% (49), Qwen3.8-27B \\
Sexual content (96) & 86.5\% (83) & 74.0\% (71) & 81.2\% (78), Qwen3Guard strict \\
Violence (96) & 77.1\% (74) & 76.0\% (73) & 58.3\% (56), Qwen3.8-27B \\
\midrule
Total (1,479) & 78.6\% (1,162) & 75.3\% (1,114) & -- \\
\midrule
\multicolumn{4}{l}{\textit{Responses}} \\
Biological weapons (97) & 87.6\% (85) & 87.6\% (85) & 53.6\% (52), Shieldstral-1.0-3B / Qwen3Guard strict \\
Chemical weapons (120) & 91.7\% (110) & 90.0\% (108) & 80.0\% (96), Qwen3Guard loose \\
Child safety (non-sexual) (79) & 94.9\% (75) & 88.6\% (70) & 43.0\% (34), Qwen3Guard strict \\
Coercive control (grooming) (95) & 89.5\% (85) & 88.4\% (84) & 54.7\% (52), Qwen3Guard strict \\
Cyber offense (75) & 97.3\% (73) & 97.3\% (73) & 77.3\% (58), Shieldstral-1.0-3B / Qwen3Guard loose \\
Fraud (75) & 100.0\% (75) & 97.3\% (73) & 78.7\% (59), Qwen3Guard loose \\
Hate \& harassment (59) & 94.9\% (56) & 94.9\% (56) & 81.4\% (48), Qwen3Guard loose \\
Illegal activities (80) & 90.0\% (72) & 85.0\% (68) & 67.5\% (54), Qwen3Guard strict \\
Privacy (65) & 95.4\% (62) & 95.4\% (62) & 64.6\% (42), Shieldstral-1.0-3B / Qwen3Guard strict \\
Radiological \& nuclear (48) & 95.8\% (46) & 93.8\% (45) & 81.2\% (39), Shieldstral-1.0-3B \\
Self-harm (74) & 93.2\% (69) & 93.2\% (69) & 51.4\% (38), GLM-5.3-Flash \\
Sex crimes (63) & 95.2\% (60) & 95.2\% (60) & 66.7\% (42), Shieldstral-1.0-3B / Qwen3Guard strict \\
Sexual content (52) & 98.1\% (51) & 94.2\% (49) & 96.2\% (50), Shieldstral-1.0-3B / Qwen3Guard strict \\
Violence (58) & 86.2\% (50) & 86.2\% (50) & 72.4\% (42), Qwen3Guard strict \\
\midrule
Total (1,040) & 93.2\% (969) & 91.5\% (952) & -- \\
\bottomrule
\end{tabular}
\end{adjustbox}
\end{table}

\FloatBarrier
\subsection{Public Benchmarks}
\label{sec:B.4}

We use four request benchmarks (WildGuardTest \citep{han2024wildguard}, Aegis v2 \citep{ghosh2025aegis2}, OpenAI Moderation \citep{markov2022holistic}, and HarmBench \citep{mazeika2024harmbench}) and six response benchmarks (WildGuardTest, BeaverTails \citep{ji2023beavertails}, Aegis v2, HarmBench, Qwen3GuardTest \citep{zhao2025qwen3guard}, and XSTestHarm \citep{rottger2023xstest}), totaling 5,582 request items and 7,650 response items.

We use the same prompting and default-decision rules as in the ISE evaluation. For each benchmark, we report F1 on three views: the whole test set with its original labels, a subset filtered for agreement with our policy, and a class-balanced version of that subset. For each view, we report the unweighted mean of per-benchmark F1 scores, separately for requests and responses, giving each benchmark equal weight regardless of its size.

DeepSeek-V4.1-Flash and GLM-5.3-Flash classify each item against our policy to construct the filtered subsets. An item is retained with its original label if at least one judge agrees with its benchmark label. Because two compared models assist in selecting these subsets, filtered outcomes are conditional on their decisions, so we also present full-test-set results. Filtering excluded 891 requests (16.0\%) and 1,242 responses (16.2\%), primarily items with harmful benchmark labels (792 requests, 1,100 responses). Per-benchmark details appear in \hyperref[tab:B.4a]{Table B.4a}.

For the class-balanced view, we upsample each filtered subset's minority label with replacement until safe and harmful items are equal. HarmBench requests have no safe items, so they lack a class-balanced view and cannot assess false positives.

\begin{table}[htbp]
\centering
\small
\phantomsection\label{tab:B.4a}
\caption*{Table B.4a. Items in each public benchmark before and after filtering. Counts are shown as whole $\rightarrow$ filtered.}
\begin{adjustbox}{max width=\linewidth}
\begin{tabular}{lcccc}
\toprule
\textbf{Benchmark} & \textbf{Items} & \textbf{Removed} & \textbf{Harmful} & \textbf{Safe} \\
\midrule
\multicolumn{5}{l}{\textit{Requests}} \\
WildGuardTest & 1,699 $\rightarrow$ 1,461 & 238 (14.0\%) & 754 $\rightarrow$ 541 & 945 $\rightarrow$ 920 \\
Aegis v2 & 1,964 $\rightarrow$ 1,573 & 391 (19.9\%) & 1,059 $\rightarrow$ 712 & 905 $\rightarrow$ 861 \\
OpenAI Moderation & 1,680 $\rightarrow$ 1,518 & 162 (9.6\%) & 522 $\rightarrow$ 390 & 1,158 $\rightarrow$ 1,128 \\
HarmBench & 239 $\rightarrow$ 139 & 100 (41.8\%) & 239 $\rightarrow$ 139 & 0 $\rightarrow$ 0 \\
\midrule
Total & 5,582 $\rightarrow$ 4,691 & 891 (16.0\%) & 2,574 $\rightarrow$ 1,782 & 3,008 $\rightarrow$ 2,909 \\
\midrule
\multicolumn{5}{l}{\textit{Responses}} \\
WildGuardTest & 1,709 $\rightarrow$ 1,607 & 102 (6.0\%) & 284 $\rightarrow$ 198 & 1,425 $\rightarrow$ 1,409 \\
BeaverTails & 3,021 $\rightarrow$ 2,346 & 675 (22.3\%) & 1,733 $\rightarrow$ 1,120 & 1,288 $\rightarrow$ 1,226 \\
Aegis v2 & 813 $\rightarrow$ 631 & 182 (22.4\%) & 394 $\rightarrow$ 226 & 419 $\rightarrow$ 405 \\
HarmBench & 602 $\rightarrow$ 524 & 78 (13.0\%) & 273 $\rightarrow$ 236 & 329 $\rightarrow$ 288 \\
Qwen3GuardTest & 1,059 $\rightarrow$ 865 & 194 (18.3\%) & 569 $\rightarrow$ 379 & 490 $\rightarrow$ 486 \\
XSTestHarm & 446 $\rightarrow$ 435 & 11 (2.5\%) & 78 $\rightarrow$ 72 & 368 $\rightarrow$ 363 \\
\midrule
Total & 7,650 $\rightarrow$ 6,408 & 1,242 (16.2\%) & 3,331 $\rightarrow$ 2,231 & 4,319 $\rightarrow$ 4,177 \\
\bottomrule
\end{tabular}
\end{adjustbox}
\end{table}

\begin{table}[htbp]
\centering
\small
\phantomsection\label{tab:B.4b}
\caption*{Table B.4b. Request F1 on public benchmarks at each model's default decision with our policy given to follow, shown as whole test set / policy-filtered / class-balanced view. HarmBench requests contain no safe items, so that row does not measure false positives or have the class-balanced view. Qwen3Guard-Gen-8B is omitted as it uses its own fixed built-in policy rather than a custom policy.}
\begin{adjustbox}{max width=\linewidth}
\begin{tabular}{lccccccc}
\toprule
\textbf{Benchmark} & \textbf{Iris Content Guard} & \textbf{GPT-OSS-Safeguard-120B (high)} & \textbf{GPT-OSS-Safeguard-20B (high)} & \textbf{DeepSeek-V4.1-Flash (effort 100)} & \textbf{GLM-5.3-Flash (max)} & \textbf{Qwen3.8-27B (xhigh)} & \textbf{Shieldstral-1.0-3B} \\
\midrule
WildGuardTest & 0.890 / 0.951 / 0.962 & 0.712 / 0.845 / 0.848 & 0.732 / 0.857 / 0.862 & 0.805 / 0.931 / 0.936 & 0.828 / 0.935 / 0.942 & 0.796 / 0.932 / 0.937 & 0.890 / 0.933 / 0.948 \\
Aegis v2 & 0.801 / 0.945 / 0.950 & 0.733 / 0.907 / 0.911 & 0.716 / 0.884 / 0.888 & 0.754 / 0.936 / 0.940 & 0.765 / 0.942 / 0.946 & 0.761 / 0.935 / 0.939 & 0.840 / 0.866 / 0.883 \\
HarmBench & 0.774 / 0.996 / -- & 0.698 / 0.955 / -- & 0.718 / 0.970 / -- & 0.749 / 0.993 / -- & 0.762 / 0.996 / -- & 0.732 / 0.985 / -- & 0.796 / 0.963 / -- \\
OpenAI Moderation & 0.798 / 0.873 / 0.936 & 0.619 / 0.743 / 0.750 & 0.720 / 0.831 / 0.852 & 0.770 / 0.900 / 0.921 & 0.763 / 0.894 / 0.918 & 0.781 / 0.890 / 0.920 & 0.750 / 0.733 / 0.886 \\
\midrule
Average & 0.816 / 0.941 / 0.949 & 0.690 / 0.862 / 0.836 & 0.722 / 0.886 / 0.867 & 0.770 / 0.940 / 0.932 & 0.779 / 0.942 / 0.935 & 0.767 / 0.935 / 0.932 & 0.819 / 0.874 / 0.906 \\
\bottomrule
\end{tabular}
\end{adjustbox}
\end{table}

\begin{table}[htbp]
\centering
\small
\phantomsection\label{tab:B.4c}
\caption*{Table B.4c. Response F1 on public benchmarks at each model's default decision with our policy given to follow, shown as whole test set / policy-filtered / class-balanced view. Qwen3Guard-Gen-8B is omitted as it uses its own fixed built-in policy rather than a custom policy.}
\begin{adjustbox}{max width=\linewidth}
\begin{tabular}{lccccccc}
\toprule
\textbf{Benchmark} & \textbf{Iris Content Guard} & \textbf{GPT-OSS-Safeguard-120B (high)} & \textbf{GPT-OSS-Safeguard-20B (high)} & \textbf{DeepSeek-V4.1-Flash (effort 100)} & \textbf{GLM-5.3-Flash (max)} & \textbf{Qwen3.8-27B (xhigh)} & \textbf{Shieldstral-1.0-3B} \\
\midrule
WildGuardTest & 0.783 / 0.892 / 0.945 & 0.673 / 0.823 / 0.834 & 0.685 / 0.826 / 0.857 & 0.735 / 0.897 / 0.933 & 0.744 / 0.893 / 0.934 & 0.742 / 0.902 / 0.923 & 0.799 / 0.876 / 0.959 \\
HarmBench & 0.839 / 0.937 / 0.942 & 0.743 / 0.852 / 0.853 & 0.762 / 0.870 / 0.872 & 0.815 / 0.938 / 0.940 & 0.838 / 0.950 / 0.954 & 0.811 / 0.923 / 0.925 & 0.855 / 0.913 / 0.924 \\
BeaverTails & 0.799 / 0.936 / 0.940 & 0.710 / 0.904 / 0.905 & 0.722 / 0.914 / 0.915 & 0.749 / 0.953 / 0.954 & 0.735 / 0.938 / 0.939 & 0.746 / 0.937 / 0.938 & 0.852 / 0.933 / 0.937 \\
XSTestHarm & 0.951 / 0.960 / 0.992 & 0.920 / 0.942 / 0.948 & 0.930 / 0.945 / 0.972 & 0.917 / 0.979 / 0.990 & 0.922 / 0.965 / 0.976 & 0.928 / 0.972 / 0.977 & 0.924 / 0.939 / 0.971 \\
Aegis v2 & 0.784 / 0.932 / 0.956 & 0.611 / 0.817 / 0.824 & 0.629 / 0.820 / 0.834 & 0.680 / 0.910 / 0.919 & 0.679 / 0.893 / 0.905 & 0.688 / 0.916 / 0.922 & 0.850 / 0.829 / 0.894 \\
Qwen3GuardTest & 0.767 / 0.942 / 0.943 & 0.689 / 0.872 / 0.873 & 0.713 / 0.887 / 0.888 & 0.771 / 0.958 / 0.959 & 0.770 / 0.948 / 0.950 & 0.769 / 0.952 / 0.952 & 0.841 / 0.936 / 0.942 \\
\midrule
Average & 0.820 / 0.933 / 0.953 & 0.724 / 0.868 / 0.873 & 0.740 / 0.877 / 0.890 & 0.778 / 0.939 / 0.949 & 0.781 / 0.931 / 0.943 & 0.781 / 0.934 / 0.940 & 0.853 / 0.904 / 0.938 \\
\bottomrule
\end{tabular}
\end{adjustbox}
\end{table}

\FloatBarrier
\section{Policy Categories}
\label{sec:C}

\begin{table}[htbp]
\centering
\small
\phantomsection\label{tab:C.1}
\caption*{Table C.1. ICG's 14-category safety policy, with the content each category treats as harmful.}
\begin{adjustbox}{max width=\linewidth}
\begin{tabular}{p{0.3\linewidth}p{0.64\linewidth}}
\toprule
\textbf{Harm Category} & \textbf{Examples of Harmful Content} \\
\midrule
Abuse Entrapment \& Coercive Control (``Grooming'') & Content that enables isolation, non-consensual surveillance, threats, manipulation, or restriction of resources to control another person or prevent them from leaving an abusive situation. \\
Biological Weapons & Content that materially enables biological harm, including acquiring, producing, enhancing, weaponizing, disseminating, or concealing dangerous biological agents or toxins. \\
Chemical Weapons & Content that materially enables chemical harm, including producing, acquiring, modifying, weaponizing, or deploying dangerous chemical agents or toxins. \\
Child Safety (Non-CSAM) & Content that enables or normalizes non-sexual harm to children, including deceptive contact, isolation, trafficking, abusive discipline, coercion, concealment of abuse, or circumvention of child-protection safeguards. \\
Cyber Offense & Content that materially enables unauthorized system access, malware development or deployment, data compromise, or cyberattacks, including operational exploitation guidance and detection or forensic evasion. \\
Fraud \& Deception & Content that enables deception for financial or material gain, including usable scam scripts, fraudulent documents, impersonation, and social engineering targeting real people or institutions. \\
Hate Speech \& Harassment & Content that attacks protected identity groups through slurs, dehumanization, harmful stereotypes, advocacy of violence, discrimination or exclusion, or denial or justification of atrocities. \\
Illegal Activities & Content that provides actionable assistance or encouragement for criminal activity, including theft, property destruction, illicit goods or controlled substances, smuggling, unlawful entry, and concealment or evasion of law enforcement. \\
Privacy Violations & Content that exposes sensitive personal identifiers not supplied by the user, invades an individual's privacy, or provides personal information that could compromise their physical, digital, or financial security. \\
Radiological and Nuclear Weapons & Content that materially enables acquiring or processing radioactive or fissile materials for harm, designing or deploying nuclear weapons or radiological devices, or repurposing medical and industrial radiation sources for malicious use. \\
Self-Harm \& Suicide & Content that enables, encourages, endorses, or compares methods of suicide, self-injury, or disordered eating, including actionable methods and lethality details. \\
Sex-Related Crimes (Non-CSAM) & Content that enables, encourages, or endorses sexual assault, coercion, harassment, trafficking, commercial sexual exploitation, or other unlawful sexual conduct, including assistance to conceal offenses or evade consequences. \\
Sexual Content (Adult, Non-CSAM) & Content that provides graphic sexual descriptions or pornography inappropriate for a general-purpose assistant, or sexualizes identifiable real people without their consent. Sexual content with ambiguous age or absent or coerced consent also crosses prohibited boundaries. \\
Violence, Violent Crimes, Threats, and Incitement & Content that enables, threatens, encourages, or glorifies physical violence, including assault, killing, torture, terrorism, extremist recruitment, and modification of civilian weapons for lethal use. \\
\bottomrule
\end{tabular}
\end{adjustbox}
\end{table}

\FloatBarrier
\end{document}